\documentclass{opt2022}

\title[Random initialisations performing above chance and how to find them]{Random initialisations performing above chance and how to find them}

\optauthor{%
\Name{Frederik Benzing\textsuperscript{\normalfont 1}} \Email{benzingf@inf.ethz.ch}\\
\Name{Simon Schug\textsuperscript{\normalfont 2}} \Email{sschug@ethz.ch}\\
\Name{Robert Meier\textsuperscript{\normalfont 1}} \Email{romeier@inf.ethz.ch}\\
\Name{Johannes {von Oswald\textsuperscript{\normalfont 1}}} \Email{voswaldj@ethz.ch}\\
\Name{Yassir Akram\textsuperscript{\normalfont 1}} \Email{yassir.akram@inf.ethz.ch}\\
\Name{Nicolas Zucchet\textsuperscript{\normalfont 1}} \Email{nzucchet@inf.ethz.ch}\\
\Name{Laurence Aitchison\textsuperscript{\normalfont 3, *}} \Email{laurence.aitchison@bristol.ac.uk}\\
\Name{Angelika Steger\textsuperscript{\normalfont 1, *}} \Email{steger@inf.ethz.ch}\\
[1em]
\addr \textsuperscript{\normalfont 1} Department of Computer Science, ETH Zurich, Switzerland \\
\addr \textsuperscript{\normalfont 2} Institute of Neuroinformatics, ETH Zurich, Switzerland \\
\addr \textsuperscript{\normalfont 3} Department of Computer Science, University of Bristol, UK \\
\addr \textsuperscript{\normalfont *} Equal contribution \\ [-4em]
}

\begin{document}

\maketitle
\begin{abstract}%
Neural networks trained with stochastic gradient descent (SGD) starting from different random initialisations typically find functionally very similar solutions, raising the question of whether there are meaningful differences between different SGD solutions.
Entezari et al.\ recently conjectured that despite different initialisations, the solutions found by SGD lie in the same loss valley after taking into account the permutation invariance of neural networks.
Concretely, they hypothesise that any two solutions found by SGD can be permuted such that the linear interpolation between their parameters forms a path without significant increases in loss.
Here, we use a simple but powerful algorithm to find such permutations that allows us to obtain direct empirical evidence that the hypothesis is true in fully connected networks. Strikingly, we find that two networks already live in the same loss valley at the time of initialisation and averaging their random, but suitably permuted initialisation performs significantly above chance. 
In contrast, for convolutional architectures, our evidence suggests that the hypothesis does not hold. 
Especially in a large learning rate regime, SGD seems to discover diverse modes.\footnote{Code available at: \url{https://github.com/freedbee/permuted_initialisations}}.
\end{abstract}


\section{Introduction}
Neural networks trained from different initialisations typically have similar decision boundaries \cite{somepalli2022can}, their predictions agree on a large fraction of data \cite{meding2021trivial} and even their internal representations are similar \cite{li2015convergent, kornblith2019similarity}. This raises the question if  all trained instances of neural networks are created equal. 
\citet{entezari2021role} recently proposed one way to formalise this idea and hypothesised that any two solutions found by stochastic gradient descent (SGD) live in the same loss valley when accounting for the permutation invariance of neural networks. More precisely, they suggest the following hypothesis.

\begin{hypothesis}[All SGD solutions are created equal, \citet{entezari2021role}]
\label{prop:sgd-equal}
For two solutions $\theta_1, \theta_2$ obtained by SGD, there is a permuted instance of the first solution $\pi(\theta_1)$ so that there is no (significant) increase in loss along the linear interpolation between $\pi(\theta_1)$ and $\theta_2$.
\end{hypothesis}

\noindent If Hypothesis~\ref{prop:sgd-equal} is true, this would greatly improve our understanding of the loss landscape of neural networks, likely impact ensemble learning \cite{garipov2018loss} as well as network sparsification \cite{frankle2020linear} and it would potentially allow to improve optimisation and generalisation \cite{izmailov2018averaging} in neural networks.
Here, we investigate this hypothesis by explicitly constructing permutations with the desired properties, leading to a range of novel findings that elucidate the structure of neural loss landscapes. 
\begin{itemize}
\item We find strong, direct evidence in support of the hypothesis for fully connected networks of moderate and large width on standard datasets. 
\item For fully connected networks, we find that an even stronger version of the hypothesis is true: The networks already live in the same loss valley at initialisation.
In particular, averaging the random initialisations $\pi(\theta_1^{\text{init}}), \theta_2^{\text{init}}$ using the permutation revealed through training, gives a network performing significantly above chance.
\item For convolutional neural networks (CNNs), our evidence suggests that there is a notable loss barrier between solutions also after accounting for permutation invariance. In particular, and interestingly, in the large learning rate regime, SGD seems to explore a genuinely diverse set of solutions.
\end{itemize}

\noindent We discuss
exciting concurrent work \cite{ainsworth2022git} as well as generally related work on permutation invariance and mode connectivity in Appendix~\ref{app:rel} .
\begin{figure}[t]
	\centering
	\includegraphics[width=0.8\textwidth]{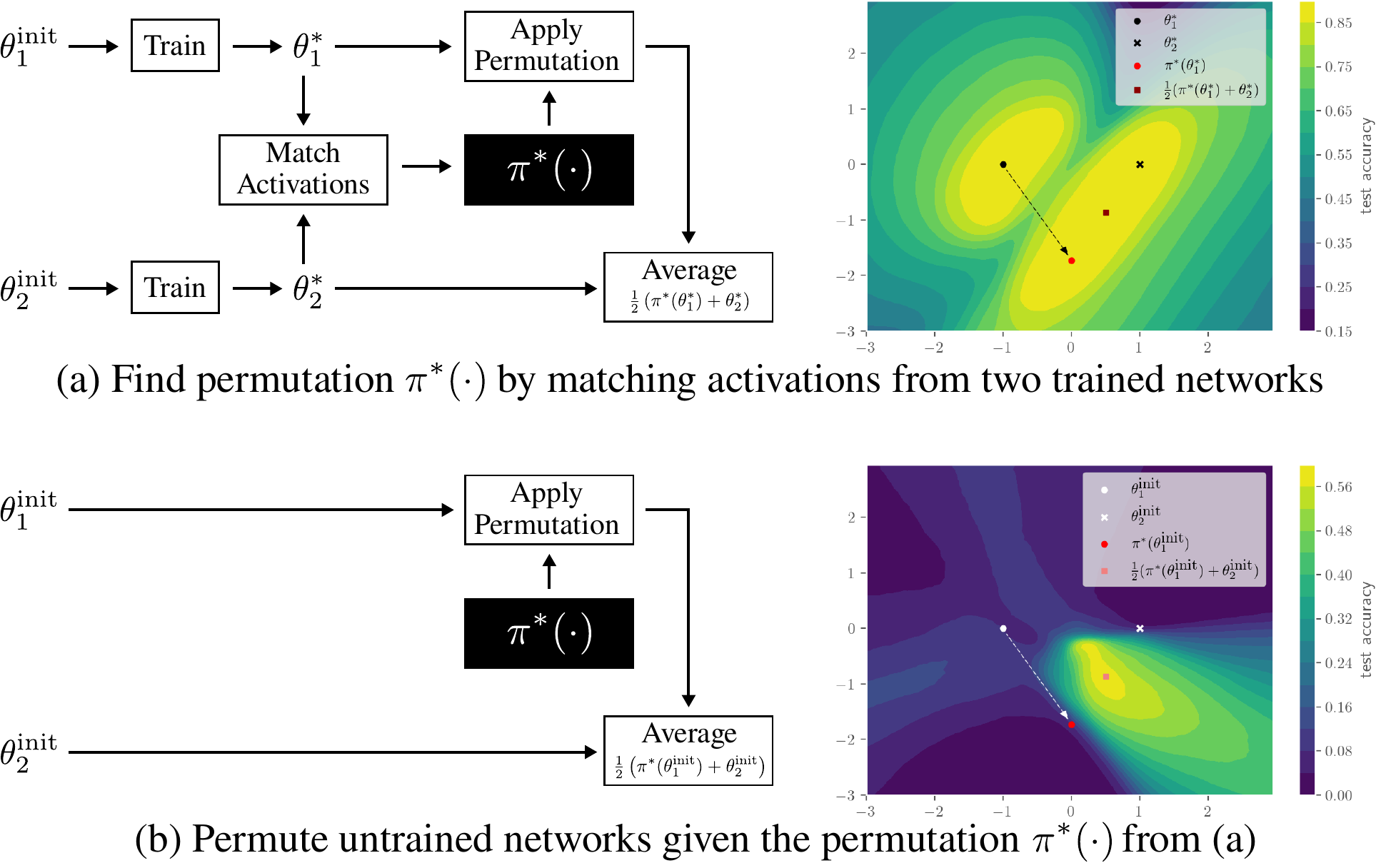}
	\caption{\textbf{Training procedure and accuracy heat map of plane spanned by three networks, $\mathbf{\theta_1, \theta_2, \pi(\theta_1)}$, for trained networks (top) and at initialisation (bottom).} After training two independently initialised networks, performance decreases when the trained weights are interpolated. This barrier vanishes after applying the permutation found by matching activations. Strikingly, when applying the permutation to the untrained initialisations, interpolating between the two results yields significantly above chance performance. Data from a fully connected network trained on Fashion MNIST.
	}
	\label{fig:graphical_abstract+heat_map}
\end{figure}

\section{Background}
We would like to evaluate whether two sets of parameters $\theta_1$, $\theta_2$ lie in the same loss valley after accounting for permutation invariance. To this end, we seek a permutation $\pi^*$ of the parameters $\theta_1$ that minimizes the loss barrier between the two sets of parameters after applying the permutation, i.e.\ we would like to solve
\begin{gather}
\min_\pi \, \mathrm{Barrier}(\pi(\theta_1), \theta_2)
\\
\mathrm{Barrier}(\theta_1, \theta_2) \coloneqq \max_{\lambda\in[0,1]} \ell\bigl(\lambda \theta_1 + (1-\lambda \theta_2)\bigr) - \Bigl( \lambda\ell(\theta_1) + (1-\lambda)\ell(\theta_2) \Bigr)
\end{gather}
While this problem is computationally intractable, a simple heuristic allows us to find good permutations in practice.
We relax the problem by considering the neuron activations of each layer separately. For each layer, we then find a permutation that makes the resulting neuron activations as similar as possible.

Formally, given a set of $d$ data points, the neuron activations of the networks parameterized by $\theta_1$ and $\theta_2$ are given by the matrices $A, B \in \mathbb{R}^{d\times n}$, where $n$ denotes the number of neurons in the hidden layer (or the number of channels for CNNs).
We define a similarity function $k(\cdot, \cdot)$ which measures for any pair of neurons $i, j$ how similar their activation vectors $A_i$, $B_j$ are and specify our surrogate objective as finding the permutation $\pi$ which maximizes the summed similarity between all matched neuron activations $s(\pi)$
\begin{align}
    s(\pi) \coloneqq \sum_{i=1}^n k(A_{\pi(i)}, B_i)
\end{align}
Maximizing $s(\pi)$ corresponds to finding a maximum matching in a bipartite graph and can be solved efficiently using the Hungarian algorithm.
We explored different variants of this idea which all performed similarly. Unless noted otherwise we use the cosine similarity as similarity function $k$ and consider the neuron activations after the non-linearity.
Finding permutations based on this idea has already been explored in \cite{li2015convergent, collier2016minimax} and re-described e.g.\ in \cite{yurochkin2019bayesian, tatro2020optimizing, singh2020model, ainsworth2022git}.






\section{Results}
\begin{figure}[t]
	{\centering
	\includegraphics[width=0.95\textwidth]{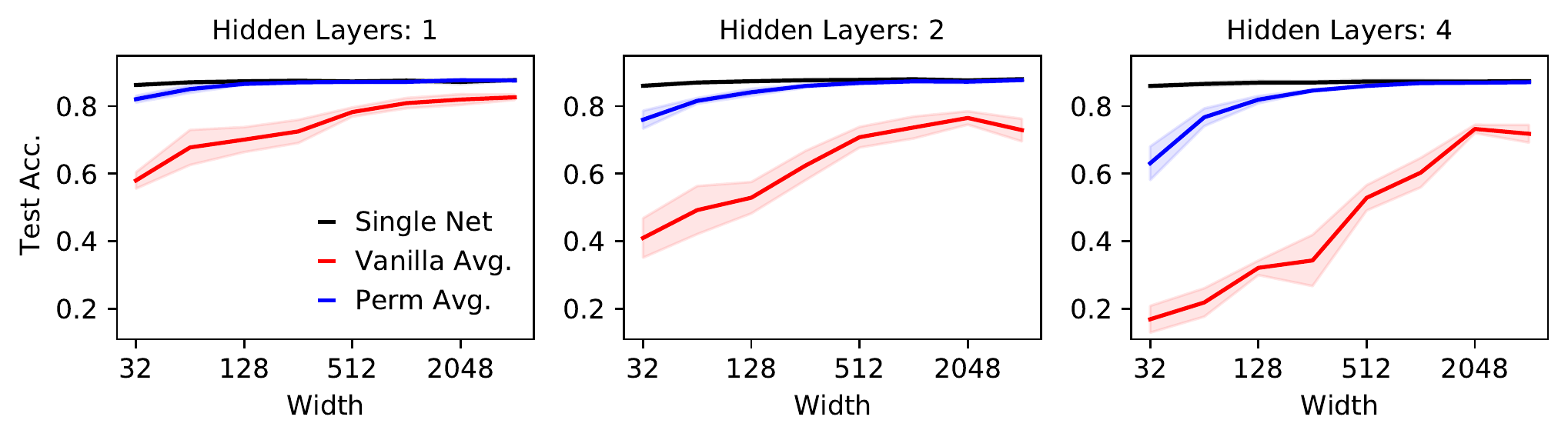}}
	{
	\centering
	\includegraphics[width=0.95\textwidth]{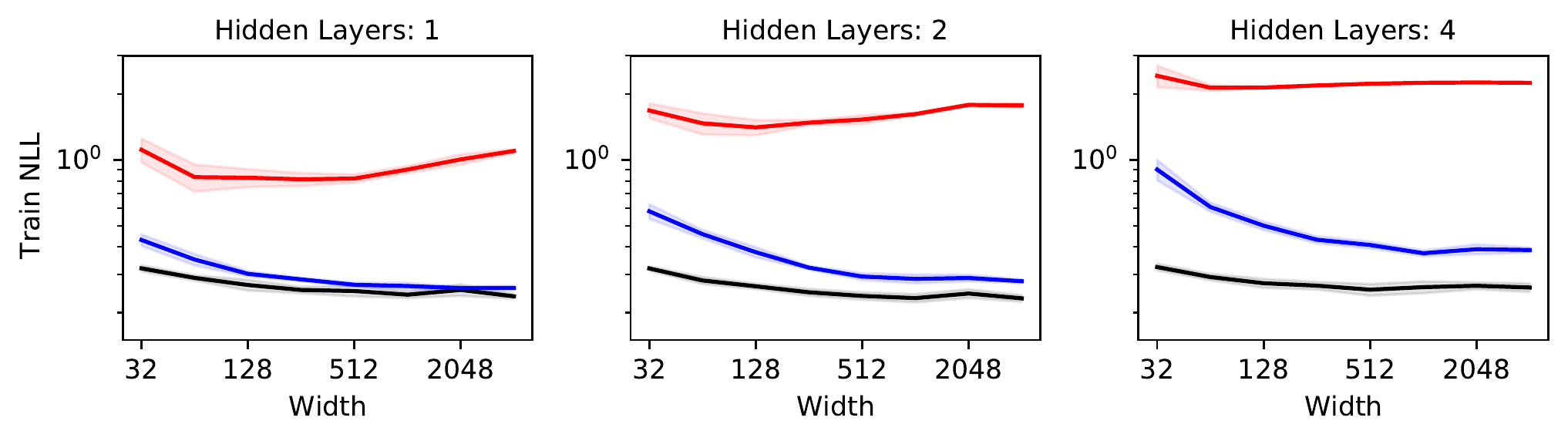}
	}
	\caption{
		\textbf{Loss barrier reduces after permutation and depends on network depth and width.}
	All networks are fully connected and trained on Fashion MNIST for five epochs with AdamW. First, we plot the performance of such a network (``Single Net'', black line). Then, we plot the performance when the parameters of two trained networks are averaged, with (``Perm Avg.'', blue) or without (``Vanilla Avg.'', red) being permuted. Thus, the loss barrier is the difference between the ``Single Net'' line (black) and the ``Perm Avg'' line (blue). Shaded region show twice the sem (approx. 95\% confidence interval) over 6 seeds.}
	\label{fig:width_depth_dependence_fashion}
\end{figure}

\subsection{Fully connected networks}
For the first set of experiments, we will focus on fully-connected networks with varying width and depth trained on MNIST \cite{lecun1998mnist}, Fashion MNIST \cite{xiao2017fashion} or CIFAR 10 \cite{krizhevsky2009learning}. In the main paper we will show results on Fashion MNIST; results on other datasets are qualitatively similar and can be found in Appendix~\ref{app:figs}.

The training and evaluation procedure is simple and outlined in Figure~\ref{fig:graphical_abstract+heat_map}(a). We train two networks independently, for 5 epochs each with AdamW \cite{kingma2014adam, loshchilov2018decoupled} with PyTorch \cite{paszke2019pytorch} default settings. We then compute the permutation making the networks' activations as similar as possible. Subsequently, we evaluate networks along the linear interpolation between the two networks, either using their unpermuted or permuted versions.


\paragraph{Trained networks lie in the same loss valley after permutation.} Figure~\ref{fig:width_depth_dependence_fashion} shows the results of these experiments. The loss barrier between solutions decreases dramatically after accounting for permutation invariance. This decrease confirms that permutation invariance is important for understanding mode connectivity.
Furthermore, by sufficiently increasing the width of the networks the loss barrier is eliminated completely by accounting for permutation invariance. Thus, in wide networks, this directly confirms the hypothesis of \citet{entezari2021role}, that there is no significant loss barrier between SGD solutions after taking permutation invariance into account.\\ 
However, in narrower networks, we did not find permutations that eliminate the loss barrier suggesting that the two solutions are genuinely different even when accounting for permutation invariance.
That said, we cannot exclude the possibility that our permutations are suboptimal, and that ``better'' permutations would eliminate the loss barrier.
Indeed, a thorough falsification of the hypothesis in narrower networks would require an exhaustive (and therefore infeasible) search over the space of permutations. 

\begin{figure}[t]
    {
	\centering
	\includegraphics[width=0.95\textwidth]{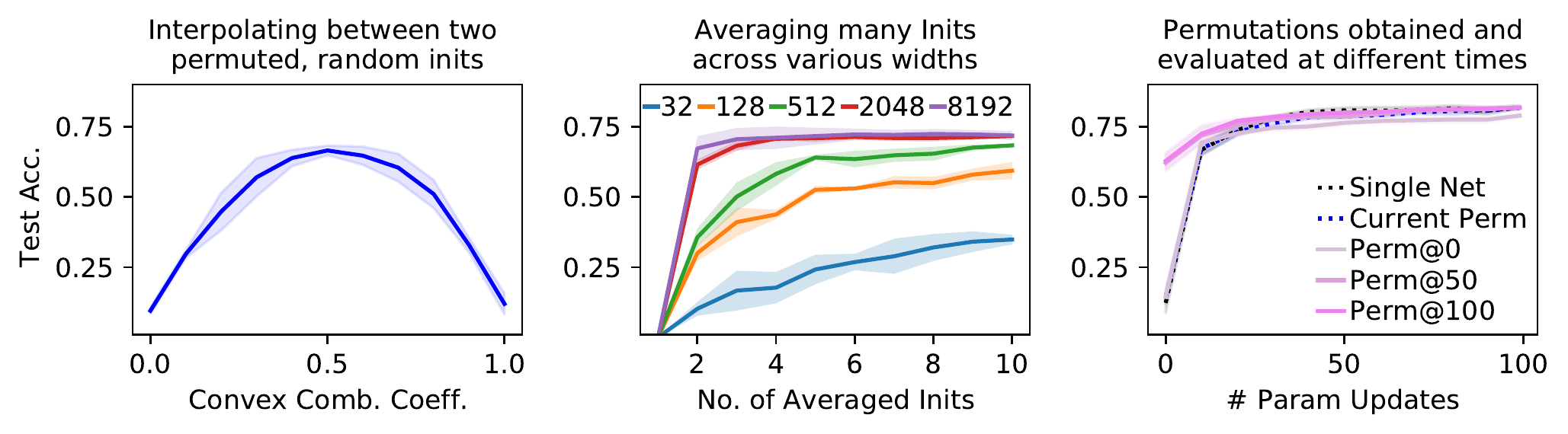}
	}
	\caption{
		\textbf{Permutations found after training project random initialisations to same loss valley.}
	\textbf{Left:} Network with two hidden layers of 2048 neurons each, trained on Fashion MNIST. Left- and rightmost point correspond to two random initialisations. The average of the two permutations performs significantly above chance, demonstrating that, after the permutation, the initialisations lie in the same loss valley. 
	\textbf{Mid:} Networks with one hidden layer and different widths. For $n>2$ networks, all networks are permuted to be similar to the first network, and then averaged with equal weights. Averaging increasing numbers of random initialisations gradually improves performance.
	\textbf{Right:} Same architecture as left panel, but permutations are computed and applied at different points at training.
	We match, permute and average the activations after each parameter update (``Current Perm'', dashed blue). 
	Additionally we apply the permutations found after $0$, $50$ respectively $100$ parameter updates at different times in training (``Perm@x'', shades of pink). Permutations that reveal that initialisations live in the same loss valley can also be found without training the networks, see Appendix~\ref{sec:permutation_without_training}.
	}
	\label{fig:init_details}
\end{figure}

\paragraph{Averaged, random initialisations perform above chance after permutations.}
We have seen that trained networks usually live in the same, or at least similar, loss valleys when accounting for permutation invariance. Next, we ask how much training is required until the networks enter a shared loss valley. 

To investigate this, we conduct experiments as outlined in Figure~\ref{fig:graphical_abstract+heat_map}(b), i.e.\ we take two random initialisations, train them with independently shuffled data, determine a permutation that makes the trained networks similar and then apply this permutation to the (untrained) initialisation.
Figure~\ref{fig:init_details}(left) shows the interpolation between those permuted initialisations and shows a rather surprising discovery: Already the initialisations live in a shared loss valley (up to a permutation), and simply averaging the permuted initialisations gives significantly above chance performance.
Figure~\ref{fig:init_details}(mid) shows that performance of these random initialisations can be further improved by averaging more than two of them. 
Figure~\ref{fig:init_details}(right) also shows that comparatively little training is required to reveal the shared loss valley between random initialisations.\\
We also note that it is possible to use the training data to find permutations that put initialisations into the same loss valley without actually training the networks, see Appendix~\ref{sec:permutation_without_training}.

\paragraph{Summary.} 
Overall, these findings give a significantly improved understanding of the loss landscape of fully-connected networks. They reveal that, up to permutation, fully-connected feedforward networks live in the same loss valley at initialisation. Moreover, even as the networks start specialising during training, the loss barrier separating any two SGD solutions remains small.

\begin{figure}[t]
    {
	\centering
	\includegraphics[width=0.95\textwidth]{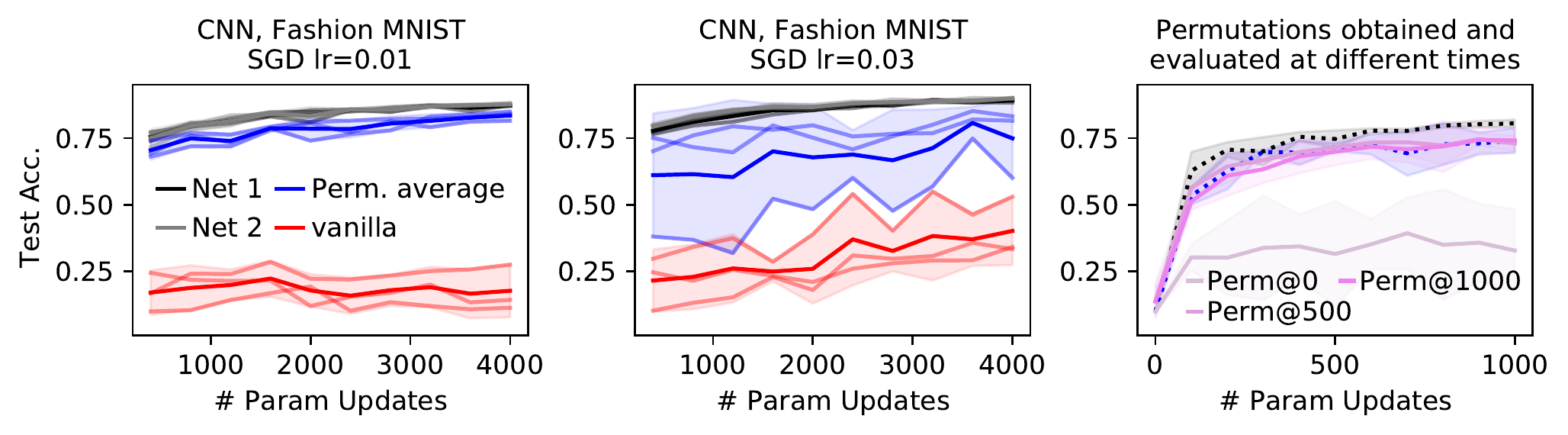}
	}
	\caption{
		\textbf{Diversity of SGD solutions depends on training regime for CNNs \& random initialisations reside in different loss valleys.}
	\textbf{Left \& Mid:} Two identical CNNs but different initialisations (Net 1 and Net 2) trained with different learning rates. With a smaller learning rate (left), the loss barrier between different solutions is very small after accounting for permutation invariance; whereas with a larger learning rate (mid) the loss barrier increases notably. This suggests that SGD explores more diverse solutions with larger learning rates. Analogous behaviour can be observed in other settings, e.g.\ for CIFAR 10 and AdamW, see Appendix~\ref{app:figs}.
	\textbf{Right:} Different permutations applied across different times, analogous to Figure~\ref{fig:init_details}~(right). This plot also shows that for CNNs the average of permuted initialisations does not perform notably above chance.}
	\label{fig:cnn}
\end{figure}
\subsection{Convolutional networks}
Next, we investigate simple CNNs purely composed of convolutional layers with no skip connections and a linear readout layer (see Appendix~\ref{app:details} for network and training details).

We again find that accounting for permutation invariance significantly reduces the loss barrier between SGD solutions, in a wide range of settings and as exemplified in Figure~\ref{fig:cnn}; however a notable barrier remained in most of our settings, including wide networks. This is also consistent with results from more complex architectures, see Fig.\ 3 in \cite{tatro2020optimizing}.

In addition, we found that the loss barrier in CNNs is significantly more sensitive to training details, like the learning rate.\footnote{As a side effect, it is harder to compare the loss barrier between architectures of different width and depth, since for each architecture the choice of training setup has a non-negligible effect.}
One example of this behaviour is shown in Figure~\ref{fig:cnn} (left\&mid), where with a larger learning rate the barrier between solutions increases. We emphasise that this behaviour can be observed across different convolutional architectures, datasets and also optimizers (AdamW as well as SGD+momentum). In all these settings, the loss barrier increases for large learning rates, which are at the edge of training stability (occasionally, runs with these high learning rates fail to converge). These results suggest that for CNNs trained with large learning rates, SGD explores diverse loss valleys, which differ also after accounting for permutation invariance.\footnote{
We did not find similar behaviour for fully connected networks, despite looking for it.}
Moreover, we note that for CNNs, different initialisations do not appear to come from a shared loss valley, see Figure~\ref{fig:cnn} (right).

\paragraph{Summary.} These results reveal two interesting differences between fully-connected and convolutional architectures. Firstly, in fully-connected networks already the initialisations live in a shared loss valley, while this does not seem to be the case for CNNs. Secondly, CNNs exhibit a richer training behaviour and in particular for large learning rates, close to training instability, they appear to explore genuinely different loss valleys, while solutions found by SGD in fully connected network are more similar.

\section*{Acknowledgements}
Simon Schug was supported by grant no. PZ00P3\_186027 from the Swiss National Science Foundation.
Robert Meier was supported by grant no. CRSII5 173721 of the Swiss National Science Foundation.
Johannes von Oswald was funded by the Swiss Data Science Center (J.v.O. P18-03).
Nicolas Zucchet and Yassir Akram were supported by ETH Research Grant no.  ETH-23 21-1. 

\clearpage 
\appendix

\section{Related work}\label{app:rel}
\subsection{Concurrent related work}
Concurrently, \citet{ainsworth2022git} explored a similar approach to investigate Hypothesis~\ref{prop:sgd-equal}. In particular, the base algorithm to find permutations is analogous to ours and that in prior work \cite{li2015convergent, collier2016minimax}. While both papers agree in their key findings, they also highlight different features. The main differences are briefly described below:\\
\citet{ainsworth2022git} present results in a ResNet18 on CIFAR 10, showing that the loss barrier also decreases (and starts to disappear) with increasing width. They also show a simple counterexample to the hypothesis for a small network with artificial data and explore merging networks trained on different subsets of the data.
In contrast, our contributions contains a systematic investigation into width- \& depth-dependence of the loss barrier in fully-connected networks and we highlight the learning rate dependence of the loss barrier in CNNs. Moreover, we showcase that in fully connected networks, already the untrained initialisations live in the same loss valley.

\subsection{Prior related work}
The question how similar different neural networks are has inspired research from a variety of angles. 
For example, \citet{li2015convergent} train independent copies of CNNs and investigate how well individual neurons and groups of neurons from one network correspond to groups of different neurons; part of their methodology also relies on computing matchings analogously to our approach.
\citet{kornblith2019similarity} investigates similarity of layer representations drawing on work from neuroscience (e.g.\ \cite{kriegeskorte2008representational}) and machine learning \cite{laakso2000content}. Their similarity measure is theoretically well-motivated and suggests that different networks learn considerably more similar representations than suggested by similarity measures of prior work \cite{raghu2017svcca, wang2018towards, morcos2018insights}.
\citet{mehrer2020individual} also study network similarity with methods from neuroscience and additionally conduct insightful experiments on how inter- or intra- class variance contribute to differences between networks.
All \citep{li2015convergent, kornblith2019similarity,mehrer2020individual} find that representations agree very strongly in early layers and decay to some extent for deeper layers. This is consistent with our findings in Figure~\ref{fig:width_depth_dependence_fashion} suggesting that deeper networks are separated by higher loss barriers.

In addition, our work is closely connected to mode connectivity. It was first shown theoretically that any two modes are connected by a (potentially complex, non-linear) path of low loss in networks with one hidden layer \citep{freeman2017topology}. This was confirmed in a range of empirical settings \citep{draxler2018essentially, garipov2018loss}.
\citet{tatro2020optimizing} show that these paths can be simplified or their loss barrier lowered when taking permutation invariance into account. They find permutations similarly as us. Thus, their work is overall similar in spirit to ours, with the difference that we investigate linear paths between modes and how architecture and training choices affect mode connectivity. 
Theoretically, \cite{brea2019weight, simsek2021geometry}  investigate the connectivity between permutations of a given mode. We note that this is different from our objective to determine whether independently obtained modes are (approximately) permutations of one another.  
Additionally, there is work developing network-/kernel-representations which are invariant to permutations and rotations \cite{aitchison2021deep}.
Linear mode connectivity has also been investigated in different contexts, e.g.\ focusing on linear connections between different points from a training trajectory of neural networks \cite{goodfellow2014qualitatively};
focussing on linear connections between networks trained on independent data samples from the same initialisation \cite{nagarajan2019uniform};
relating linear mode connectivity to flatness and generalisation \cite{izmailov2018averaging}, and relating it to sparsity and the lottery ticket hypothesis \cite{frankle2018lottery, frankle2020linear}.



\section{Finding good permutations at initialisation without training}
\label{sec:permutation_without_training}
In the main paper we described how a permutation obtained by training networks and matching their features leads to averaged initialisations that perform above chance. 
Here, we briefly note that such permutations can also be obtained without training the networks and describe how this can be done.

As a first observation note that for our desired permutation $\pi$, the loss decreases when we start at $\theta_2$ and move towards $\pi(\theta_1)$, i.e.\ in the direction of $v_\pi=\pi(\theta_1) - \theta_2$. This suggests that the direction $v_\pi$ should be (negatively) aligned with the gradient at $\theta_2$. Thus, we can try to find permutations that minimize the dot product $v_\pi \cdot \frac{\partial L(\theta)}{\partial \theta}|_{\theta=\theta_2}$. This can be formalised as a minimum matching problem again and thus be solved with the Hungarian algorithm. \\
The idea described above does not yet give good permutations and we need two additional ideas. 
\begin{itemize}
\item
First, and most importantly, we replace the vanilla gradient $\frac{\partial L(\theta)}{\partial \theta}|_{\theta=\theta_2}$ by the update direction of a more powerful optimizer. We used FOOF \cite{benzing2022gradient}, which was inspired by obtaining a better understanding of KFAC \cite{martens2015optimizing, desjardins2015natural}.
\item 
Second, we symmetrise the loss. I.e.\ (using the vanilla gradient rather than more advanced optimizers for notational simplicity) we minimize
\begin{align}
    \min_\pi\;
    (\pi(\theta_1)-\theta_2) \cdot \frac{\partial L(\theta)}{\partial \theta}|_{\theta=\theta_2} 
    + 
    (\pi^{-1}(\theta_2)-\theta_1) \cdot \frac{\partial L(\theta)}{\partial \theta}|_{\theta=\theta_1} 
\end{align}
\end{itemize}
We note that this gives permutations with very similar performance (at initialisation) as the permutations described in the main text.
Moreover, in the case of CNNs this algorithm also does not find permutations which put different initialisations in the same loss valley, further supporting our finding from the main paper that CNN initialisations live in distinct loss valleys.

\section{Finding permutations for second-order Taylor expansions around the loss}\label{sec:high_order_expansion}
The algorithm described in the preceding section (almost) amounts to finding a permutation that minimizes a first-order Taylor expansion of the loss around $\theta_2$ evaluated at $\frac12\pi(\theta_1) + \frac12\theta_2$.\footnote{We choose the midpoint between $\pi(\theta_1)$ and $\theta_2$ here since empirically, this is a very good approximation of the point along the linear interpolation that maximizes the loss barrier.}
Thus or otherwise motivated, one might be tempted to find permutations that minimize higher-order Taylor expansions of the loss around $\theta_2$ evaluated at $\frac12\theta_2+\frac12\pi(\theta_1)$ both for initialisations or trained networks $\theta_2$.\\ 
If one uses KFAC \cite{martens2015optimizing} to obtain a computationally tractable approximation of the second-order expansion of the loss around $\theta_2$, and then computes the permutations sequentially layer-by-layer, the minimization problem to find good permutations turns out to be equivalent to the quadratic assignment problem (and thus NP hard). We experimented with convex relaxations of this problem over the Birkhoff polytope but failed to find better permutations than with the simpler method described in the main paper.

\section{Variants of feature matching and interpolation}
We experimented with different similarity functions $k(\cdot,\cdot)$ for measuring the similarity between neuron activations, in particular $\ell_2$-distance and cosine similarity. We also investigated extracting neuron activations both before and after the non-linearity. Broadly speaking, all variants performed very similarly.
In addition, we experimented with combining the permutations with learned / non-learned ``small'' rotations. This often decreased the loss barrier slightly, but did not affect results qualitatively. 
Moreover, we tried interpolating between each layers' weights using a rotation, rather than a linear path, but found that this significantly increased the loss barrier.

Finally, we tried matching weights instead of neuron activations. Concretely, rather than finding a permutation that maximizes similarity between activation vectors, we found a permutation that maximizes similarity between weight vectors. 
In principle, for every layer except the first and the last, this requires computing two permutations. To circumvent this problem, 
we simply started with the first layer, and moved upwards step by step.
Matching weights performed slightly worse than matching activations, i.e. the loss barrier increased.


\section{Recap of permutation invariance}
It is well known that neural networks are permutation invariant. To make this term precise, consider a fully connected network consisting of a sequence of weight matrices $W_1, \ldots, W_\ell$, where $W_i$ has dimensions $n_{i-1} \times n_i$, and a pointwise non-linearity $\sigma$ (for simplicity of notation, we assume that each layer is followed by the same non-linearity, but this assumption can easily be dropped). For a given input $X$, the output of the $i$-th layer $A_{i}$ is defined iteratively as
\begin{align}
A_0 &= X \\ 
A_i &= \sigma(W_{i} A_{i-1}) \quad\text{for } 1\leq i \leq \ell
\end{align}
and the output of the neural network is simply the output of the last layer $A_\ell$.

If we are given a sequence of permutation matrices $P_0, \ldots, P_{\ell}$, where $P_i$ has dimensions $n_i \times n_i$ and where the first and the last permutation matrix $P_0, P_\ell$ are the identity, then the permuted neural network consisting of the weight matrices 
$$\tilde{W}_i = P_i^T W_i P_{i-1}$$ computes precisely the same output as the network consisting of weight matrices $W_i$ (since non-linearity and permutation commute and since $P_i^T P_i = \mathrm{Id}$). Thus the input-output function specified by the network is invariant to permutations as described above.

While we used fully connected networks above, permutation invariance straightforwardly applies for example to convolutional networks, where channels can be permuted. 


\section{Experimental details}\label{app:details}
The term ``SGD'' in the paper usually refers to SGD with the usual 0.9 heavy-ball momentum, but we obtained qualitatively similar results for SGD without momentum.
We used a batch size of 100 in all experiments, but we found similar results across a range of batch sizes. 
The permutations are usually computed using the activations of 10000 images, and we verified that the quality of permutations saturates already for smaller numbers of images.
We used Kaiming-He init and ReLU nonlinearities for intermediate layers. The final layer is followed by a softmax, in combination with the cross-entropy loss. 


\subsection{Fully connected networks}
Networks presented here are trained for 5 epochs with AdamW PyTorch \cite{paszke2019pytorch} default settings, except Figure~\ref{fig:init_details} mid-panel, where permutations were obtained by training for one epoch. We note that for narrow networks, permutations that obtain even better performance when used to average initialisations can be obtained by training shorter. 

\subsection{CNNs}
All CNN architectures use a sequence of convolutional filters and 2x2 average pooling layers, based on standard CNN design practices. These layers are finally followed by a global average pooling layer and one fully connected layer.

Based on this general design, we trained two kinds of CNN architectures, 
one of type 
$$k - k - \text{avg} - 2k - 2k$$
where ``$\text{avg}$'' denotes a 2x2 average pooling layer and the remaining numbers denote the number of (hidden) channels in the convolutional layers; and another one of type:
$$k - \text{avg} - 2k - \text{avg} - 4k.$$
We used $k=32, 64, 128, 256$

\begin{figure}[h]
    {
	\begin{center}
	\includegraphics[width=0.6\textwidth]{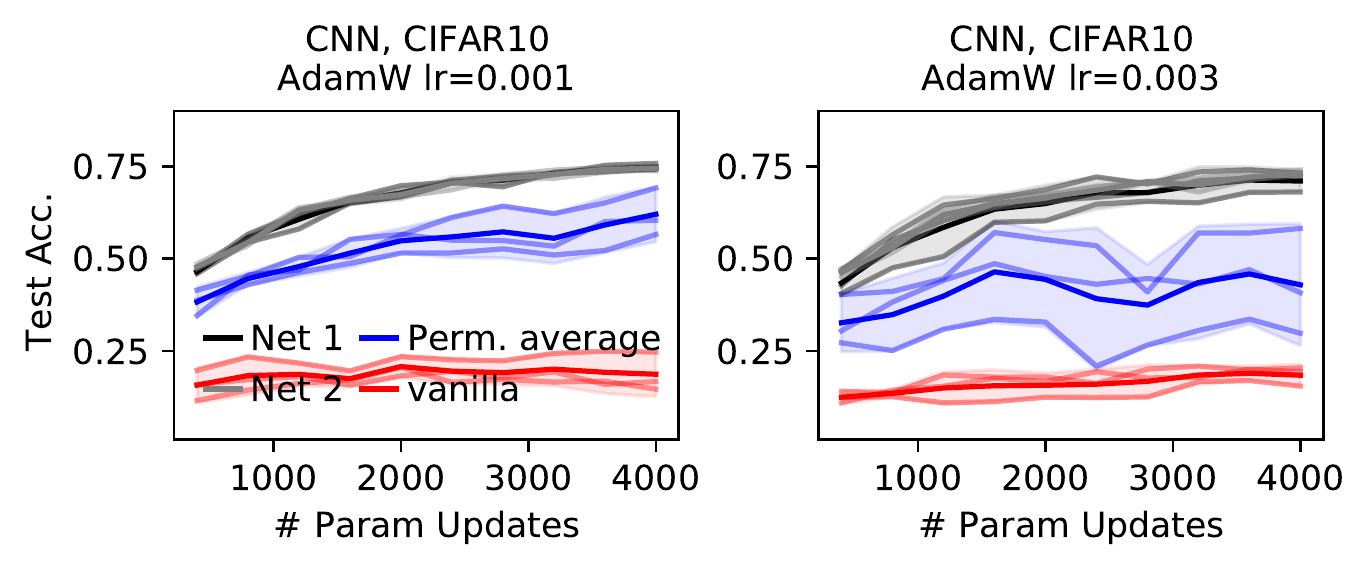}
	\end{center}
	}
	\caption{Analogous to Figure~\ref{fig:cnn}, but on CIFAR10 with AdamW.
	}
	\label{fig:cnn_cifar}
\end{figure}

The CNN in Figure~\ref{fig:cnn} is of type $128-128-\text{avg}-256-256$, but we observed similar behaviour across different optimizers, architectures and datasets.  One more example for CIFAR10, AdamW and type $64-64-\text{avg}-128-128$ is shown in Figure~\ref{fig:cnn_cifar}.

\clearpage
\section{Additional figures}\label{app:figs}
\begin{figure}[h]
	{\centering
	\includegraphics[width=0.45\textwidth]{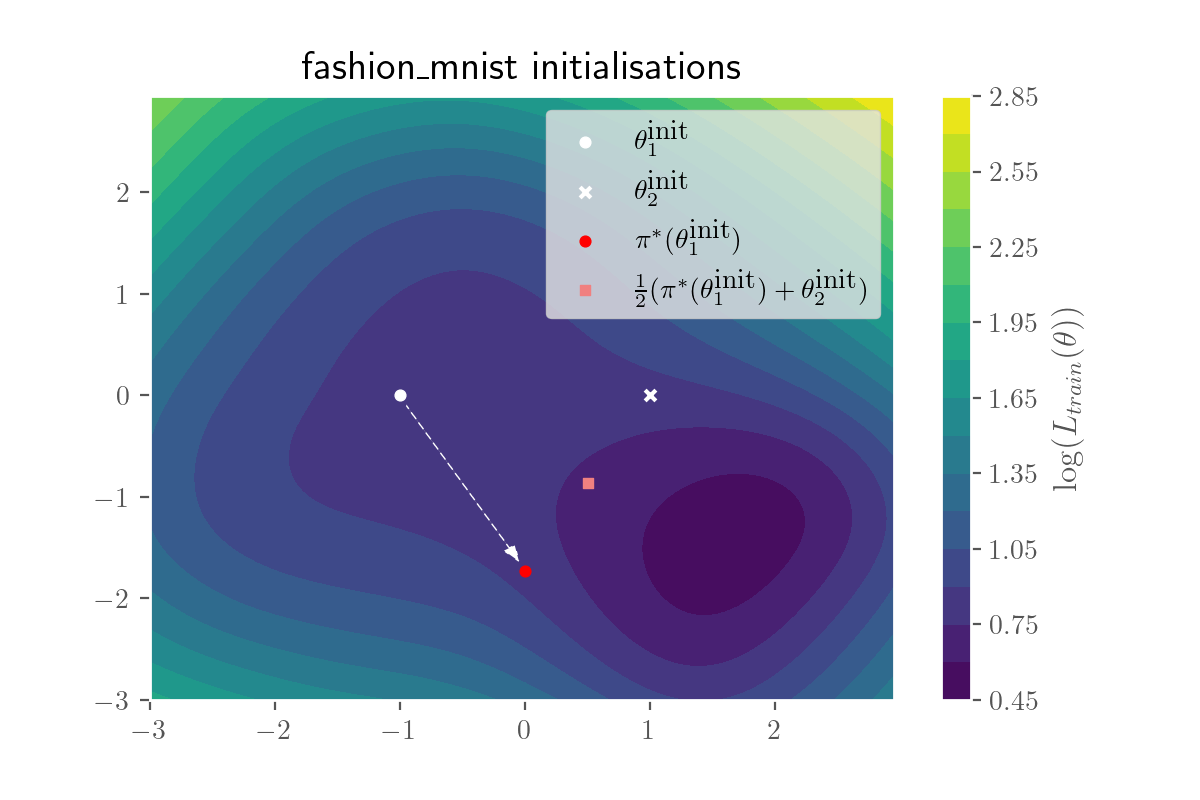}}
	{
	\centering
	\includegraphics[width=0.45\textwidth]{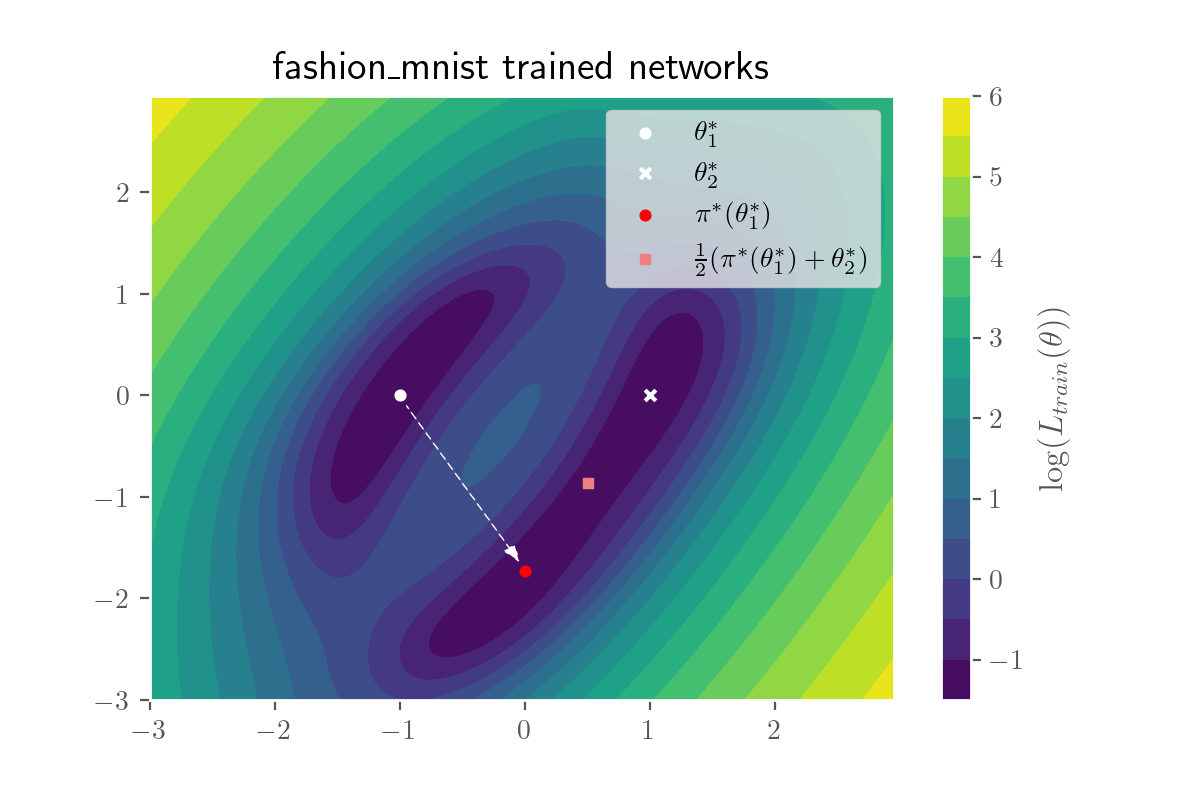}
	}
	\caption{
		Corresponding Train Loss Heat Map to the right part of Figure~\ref{fig:graphical_abstract+heat_map}.}
	\label{fig:accuracy_heatmap}
\end{figure}

\begin{figure}[h]
	{\centering
	\includegraphics[width=0.95\textwidth]{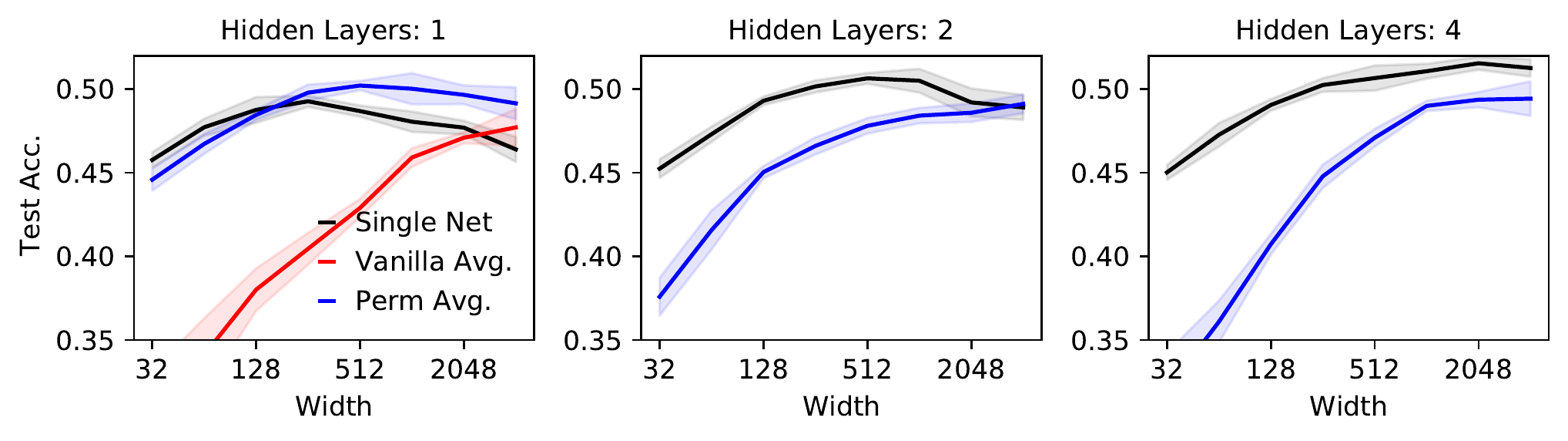}}
	{
	\centering
	\includegraphics[width=0.95\textwidth]{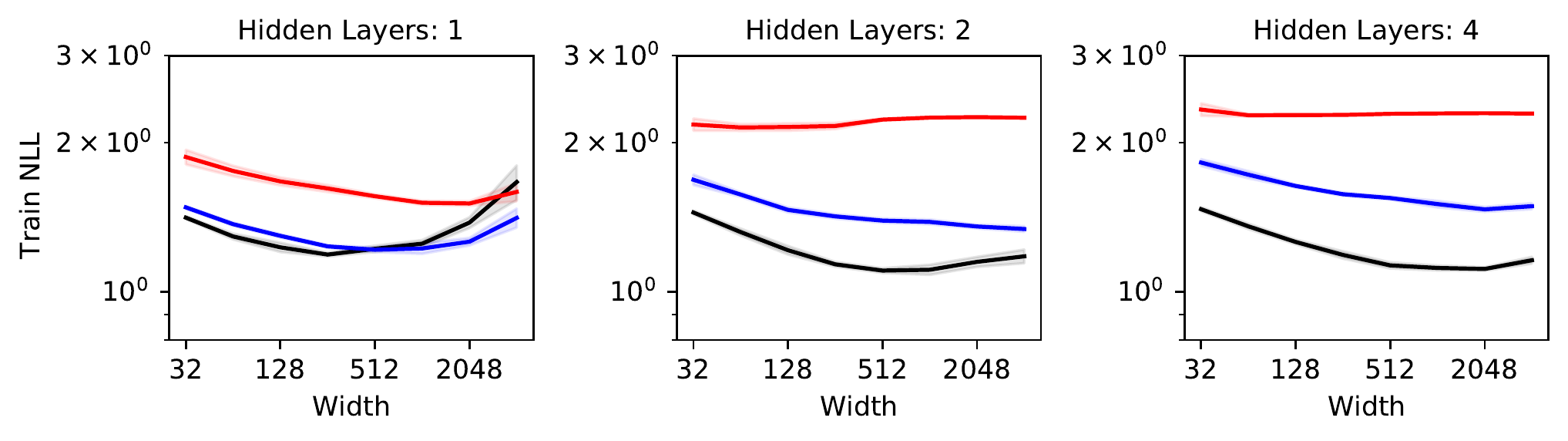}
	}
	\caption{
		Analogous to Figure~\ref{fig:width_depth_dependence_fashion}, but on CIFAR 10.\\
		Loss increases at very high width, because in this case lower learning rates would be optimal, but we stick to AdamW default settings.}
	\label{fig:width_depth_dependence_cifar}
\end{figure}

\begin{figure}[h]
	{\centering
	\includegraphics[width=0.95\textwidth]{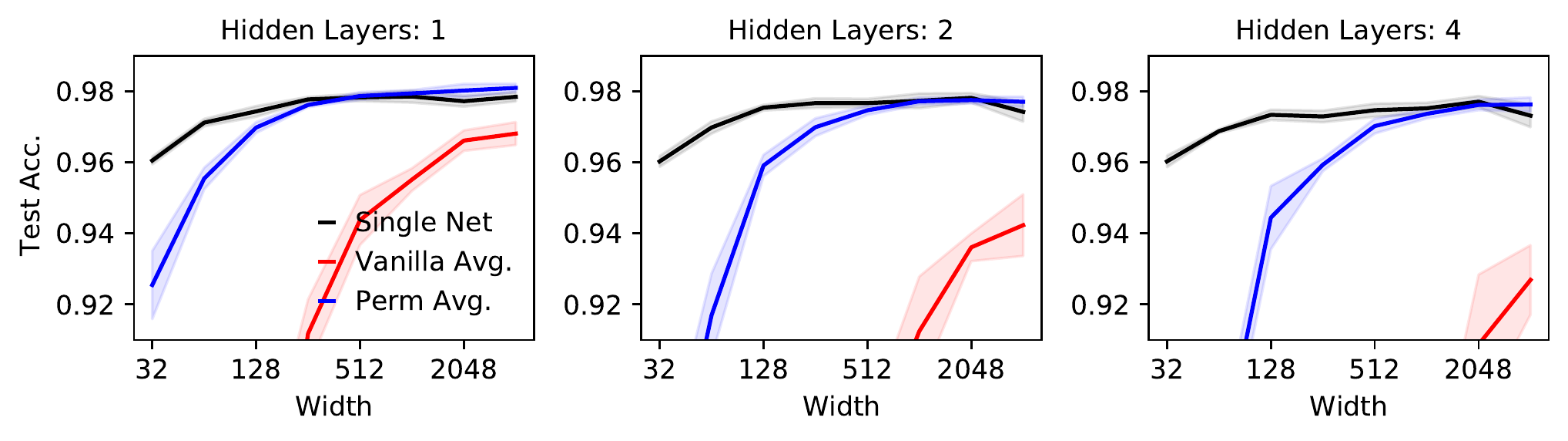}}
	{
	\centering
	\includegraphics[width=0.95\textwidth]{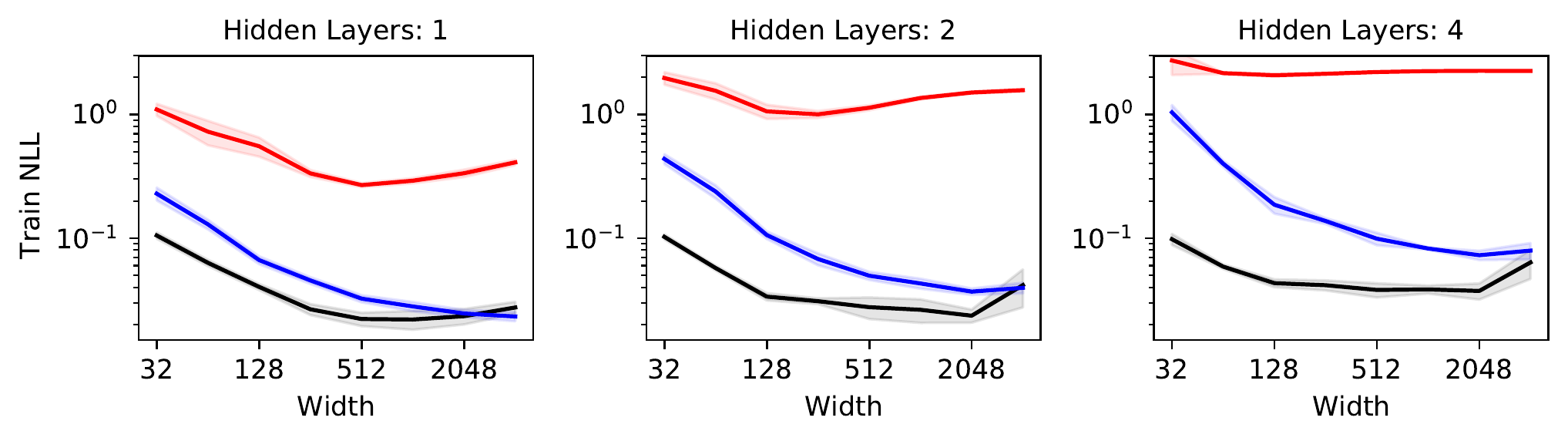}
	}
	\caption{
		Analogous to Figure~\ref{fig:width_depth_dependence_fashion}, but on MNIST.\\
		Loss increases at very high width, because in this case lower learning rates would be optimal, but we stick to AdamW default settings.}
	\label{fig:width_depth_dependence_mnist}
\end{figure}

\clearpage
\begin{figure}[h]
    {
	\centering
	\includegraphics[width=0.95\textwidth]{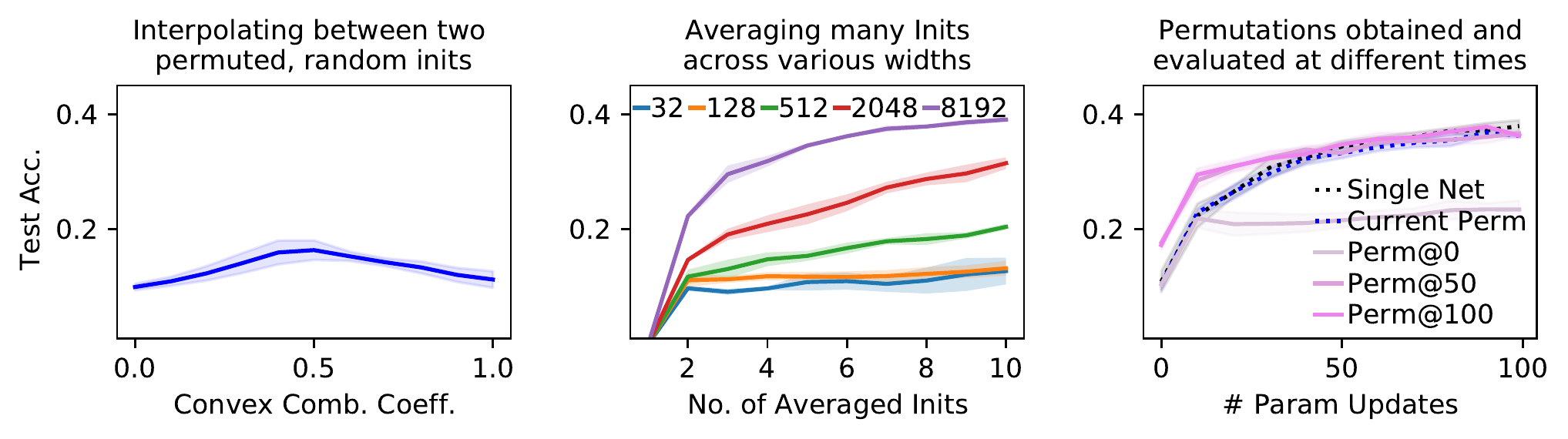}
	}
	\caption{Analogous to Figure~\ref{fig:init_details}, but on CIFAR 10.}
	\label{fig:init_details_cifar}
\end{figure}

\begin{figure}[h]
    {
	\centering
	\includegraphics[width=0.95\textwidth]{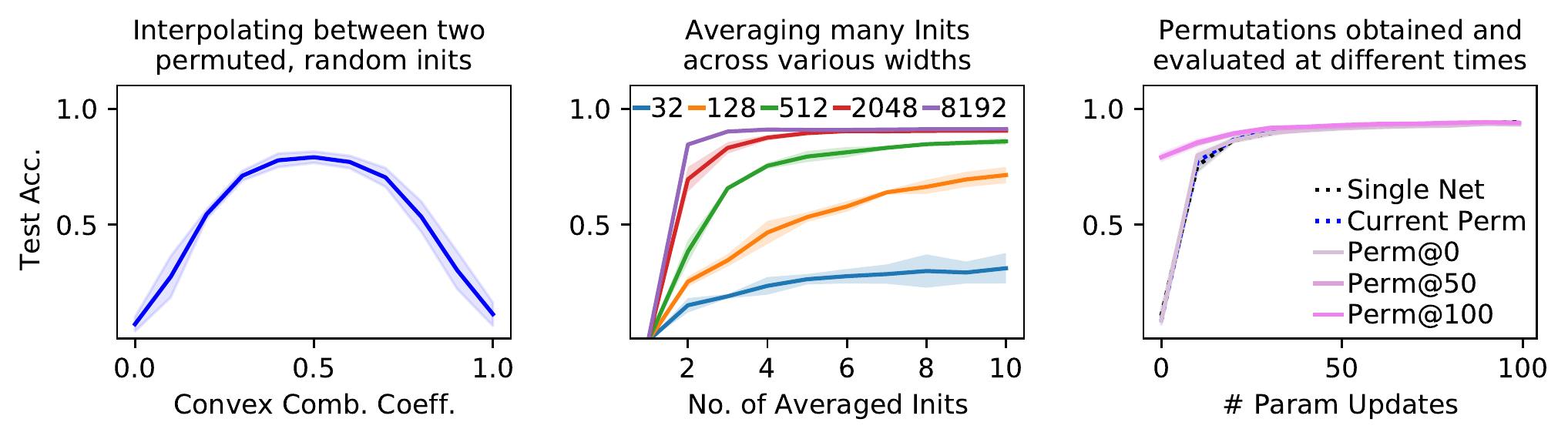}
	}
	\caption{Analogous to Figure~\ref{fig:init_details}, but on MNIST.}
	\label{fig:init_details_mnist}
\end{figure}

\begin{figure}[h]
        {\centering
	\includegraphics[width=0.49\textwidth]{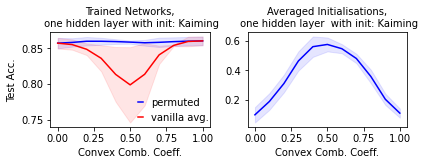}}
	{
	\centering
	\includegraphics[width=0.49\textwidth]{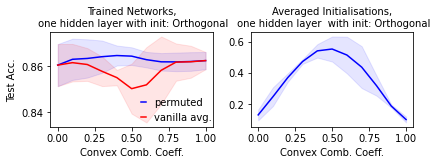}
	}
	{\centering
	\includegraphics[width=0.49\textwidth]{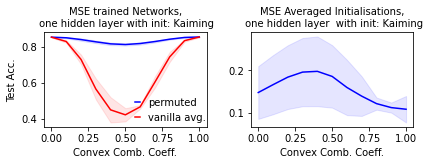}}
	{
	\centering
	\includegraphics[width=0.49\textwidth]{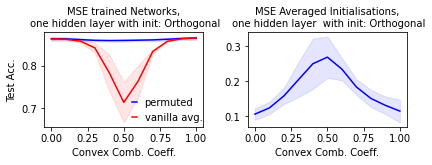}
	}
	\caption{Additional results on Fashion MNIST when varying the initialization scheme and loss function. \textit{Top row}: The effect of vanilla and permuted averaging, at initialization and after training, for two different initialization schemes. We test Kaiming-He (as in the main text) \cite{kaiming}, Xavier (not shown) and Orthogonal  \cite{DBLP:journals/corr/SaxeMG13} initialization and find that the mode connectivity remains linear for the different initializations. Behaviour with Xavier initialization is very similar to the Orthogonal one and therefore omitted. \textit{Bottom row}: When turning to the mean squared error (MSE) loss instead of the cross-entropy loss, we observe worse performance for vanilla averaging before and after training, for the two initializations, while the networks stay linearly connected after permuting. \label{fig:additional_fashion_mnist}} 
\end{figure}
\clearpage
\bibliography{references}
\end{document}